# Ferrofluidic Manipulator: Automatic Manipulation of Non-magnetic Microparticles at the Air-Ferrofluid Interface

Zoran Cenev[†], P. A. Diluka Harischandra[†], Seppo Nurmi, Mika Latikka, Ville Hynninen, Robin H. A. Ras, Jaakko V. I. Timonen and Quan Zhou[*]

*Abstract*— Manipulation of small-scale matter is a fundamental topic in micro- and nanorobotics. Numerous magnetic robotic systems have been developed for the manipulation of microparticles in an ambient environment, liquid as well as on the air-liquid interface. These systems move intrinsically magnetic or magnetically tagged objects by inducing a magnetic torque or force. However, most of the materials found in nature are non-magnetic. Here, we report a ferrofluidic manipulator for automatic two-dimensional manipulation of non-magnetic objects floating on top of a ferrofluid. The manipulation system employs eight centimeter-scale solenoids, which can move non-magnetic particles by deforming the air-ferrofluid interface. Using linear programming, we can control the motion of the non-magnetic particles with a predefined trajectory of a line, square, and circle with a precision of 25.1±19.5 μm, 34.4±28.4 μm and 33.4±26.6 μm, respectively. The ferrofluidic manipulator is versatile with the materials and the shapes of the objects under manipulation. We have successfully manipulated particles made of polyethylene, polystyrene, a silicon chip, and poppy and sesame seeds. This work shows a promising venue for the manipulation of living and non-living matter at the air-liquid interface.

## I. INTRODUCTION

MAGNETIC field-driven manipulation of miniaturized robotic agents has attracted great research attention with a wide range of applications. Impressive results have been achieved during the last two decades ranging from contactless ocular surgery [1], targeted drug delivery [2]–[4], in-vitro diagnosis [5], endoscopy [6], minimally invasive surgery [7], and environmental remediation [8], [9] to name a few. Magnetic fields have been widely used in robotic micromanipulation due to their transparency to human and animal tissues, and their capacity to wirelessly address micro-objects. It has been shown that magnetic-driven microrobotic systems are capable of performing independent control of multiple magnetic agents in 2D [10] and 3D [11], selective manipulation and extraction [12], [13], motion and patterning of swarms [14], [15], as well as carrying out targeted gene delivery [16] and climbing on liquid menisci [17], among a remaining plethora of other reported capabilities [18]–[22].

However, most of the magnetic field-driven manipulation methods require that the object under manipulation features paramagnetic or ferromagnetic properties. To manipulate non-magnetic objects, intermediate magnetic tools such as magnetic pushers [23], [24], grippers [25], swarm of magnetic agents [14], [15], and magnetic agent-induced fluid flow [26] have been employed. A magnetic spray has also been used for magnetic tagging of non-magnetic objects to enable magnetic manipulation [27]. Manipulation of non-magnetic objects on the air-liquid interfaces has also been previously done using other methods, e.g. mechanical contact [28] and thermocapillary convention [29].

Recently, magnetic liquids such as paramagnetic solutions and ferrofluids have found applications in microrobotics. For example, ferrofluidic droplets can act as shape-programmable magnetic miniature soft robots for splitting, regeneration, and navigation through small channels [30], [31]. The shape of magnetic liquids can be altered by an external magnetic field. Earlier work has reported liquid marbles can be pushed on an air-magnetic liquid interface [32]. We have recently reported the underlying physical mechanism of displacing and trapping micro- and millimeter-sized particles at the air-paramagnetic liquid interface [33]. We have shown that non-magnetic particles besides being pushed, can also be pulled, and eventually trapped at the air-paramagnetic liquid interface.

This work was supported in part by the Academy of Finland (projects #295006, #296250 and #317018).
[†]*Zoran Cenev and P. A. Diluka Harischandra are co-first authors and have contributed equally to this work.*
[*]*Corresponding author: quan.zhou@aalto.fi*

Zoran Cenev, P. A. Diluka Harischandra and Quan Zhou are with Department of Electrical Engineering and Automation, School of Electrical Engineering, Aalto University, Maarintie 8, 02150 Espoo, Finland, e-mail (zoran.cenev@aalto.fi, diluka.harischandra@aalto.fi and quan.zhou@aalto.fi).

Seppo Nurmi is with Department of Mechanical Engineering, School of Engineering, Aalto University, Otakaari 4 (K3), 02150 Espoo, Finland (e-mail: seppo.nurmi@aalto.fi).

Zoran Cenev, Mika Latikka, Ville Hynninen, Robin H. A.Ras and Jaakko V. I. Timonen are with Department of Applied Physics, School of Science, Aalto University, Puumiehenkuja 2, 02150 Espoo, Finland (e-mail: mika.latikka@aalto.fi; ville.hynninen@aalto.fi; robin.ras@aalto.fi; jaakko.timonen@aalto.fi).

Robin H. A. Ras is with Department of Bioproducts and Biosystems, Aalto University, P.O. Box 16000, 02150 Espoo, Finland

This article has supplementary downloadable multimedia material available.

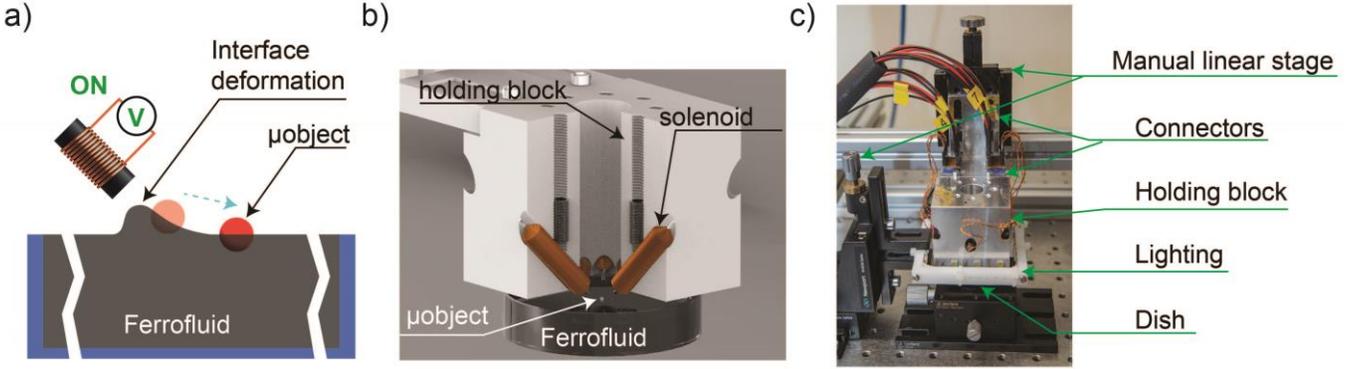

Fig. 1. Ferrofluid manipulator: the concept and interacting components for manipulating small-scale non-magnetic objects at the air-ferrofluid interface. Illustration not to scale. **a)** Concept of two-dimensional manipulation of a non-magnetic object: e.g. polyethylene particle **b)** Computer-aided design (CAD) rendered image of the mechanical assembly consisting of solenoids within a holding block located on top of a dish containing the ferrofluid and a particle at the air-ferrofluid interface. **c)** A photograph of the ferrofluidic manipulator showing the aluminum holder with the solenoids, a ring LED light, and a petri dish containing a ferrofluid sitting on a 2D goniometer stage. The manipulation platform resides on an anti-vibration table.

Here we propose a ferrofluidic manipulator that can automatically manipulate non-magnetic objects by controlling the curvature of the air-ferrofluid interface using eight solenoids. The ferrofluidic manipulator can manipulate non-magnetic particles on the air-ferromagnetic interface along predefined trajectories, without contact with a solid tool, moving magnetic agent, coating the objects, or heating. The remaining part of the paper is structured as follows: Section II describes the working principle and the mechatronic design of the manipulator. Section III reports the preparation of the ferrofluid and the types of objects used for manipulation. The open-loop performance of the ferrofluidic manipulator is experimentally characterized and described in Section IV. Section V explains a path-following algorithm based on linear programming. Finally, Section VI discusses the results and concludes the paper. The extended technical details are given in the Appendix.

## II. System Overview

### A. Concept and working principle

Fig. 1(a) illustrates the manipulation concept of the ferrofluidic manipulator. A non-magnetic micro-object with a commensurate density to the one of the ferrofluid is placed on the air-ferrofluid interface. By applying a magnetic field to the air-magnetic liquid interface, the interface deforms (into a bump) and the particle is pushed away from the interface deformation as well as from the source of the magnetic field as shown in Fig. 1(a). The height of the interface deformation (the bump) is depending on the strength of the magnetic field acting on the ferrofluidic surface, and in our manipulator, it is in the order of hundreds of micrometers. By superimposing the magnetic fields from different magnetic sources, micro- and millimeter-scale non-magnetic objects such as particles, chips, seeds can be manipulated in a desired trajectory.

In general, the motion of a particle pinned at the air-ferrofluid interface induced by a non-uniform magnetic field represents an energy minimization problem, where the particle will move towards the regions with minimum energy. In an axis-symmetric configuration, the total energy has the following form [33]:

$$E = E_0 + m_e g(u(\rho) + l^2 H) - (\chi_L V_{imm} - \chi_p V_p)B^2 \frac{1}{\mu_0} \quad (1)$$

where $E_0$ is surface adsorption energy, $m_e$ is the effective mass, $g$ is gravitational acceleration, $\rho$ is the distance along the horizontal axis for the polar coordinate system and $u(\rho)$ is the interface deformation caused by the magnetic field $B$, $l$ is the capillary length of the magnetic liquid, $H$ is the mean curvature of the interface deformation, $\chi_L$ and $\chi_p$ are magnetic susceptibility of the liquid and the particle, respectively, $V_{imm}$ is the immersion volume and $V_p$ the volume of the particle, and $\mu_0$ is the permeability of free space.

In Eq. (1), the first term is a constant, the second term quantifies the contributions from the gravitational and the capillary energies, whereas the third term quantifies the magnetic energy. The force acting on a particle can be derived as a negative derivative ($F = -dE/d\rho$) from the total energy expressed in Eq. (1). Note that conversion to Cartesian coordinates is necessary to include the contribution from the inclined magnetic field sources. The actual influence of the parameters such as the overall magnetic field, the distance of the particle to the solenoid(s), and the size of the particle are very challenging to be quantified theoretically since the gravitational and capillary contributions will depend on the non-linear properties of the interface deformation and its derivatives. Therefore, some relations are experimentally determined in Section III.

### B. System components and mechatronic integration

Fig. 1(b) illustrates the computer-aided design of the mechanical assembly consisting of eight centimeter-scale solenoids within a holding block located on the top of a petri dish (diameter: ~57 mm) containing the ferrofluid and a particle at the air-ferrofluid interface. Fig. 1(c), shows the major components. The workspace is bounded by the solenoids in a circular arrangement with an offset of 45° from the horizontal plane and radially from each other as shown in Fig.1(b). Manipulated objects are placed on the air-ferrofluid interface prior to any manipulation experiment. The manipulation workspace was imaged by a camera (Point Gray GS3-U3-41C6C-C, Flir Systems Inc., Canada) with a zoom macro lens (Zoom Macro 7000, Navitar, USA) placed on the top of the solenoids holding block providing visual information. The solenoids were controlled in a digital ON/OFF manner. The control signal was generated by a computer and transmitted to the custom-made electronic circuitry via a digital input/output board (NI DIO 6501, National Instruments, USA). The control software was written in C++ (Visual C++ version 16.0, Microsoft, USA). Image and video and data recording were performed using in-house developed software. Image processing algorithms were developed using ViSP (Visual Servoing Platform) library [34]. The workspace was illuminated with a light-emitting-diode (LED) light stripe (Velleman, Belgium).

The solenoids have a ~4 cm metal core made of one-millimeter-thick martensitic stainless-steel wire (AISI 420), wrapped with copper wire (SWG 27, PKC Group, Finland) in 6 layers with 300 to 400 coil turns. Each coil has a resistance of ~0.5 Ω and inductance of 200 to 250 mH. The frontal end of each solenoid was grinded flat. The long solenoids had neck length of 4 to 6 mm and short solenoids of 1 to 2 mm. The magnetic properties of the solenoids are elaborated in Appendix A. The tips of the solenoids are positioned 0.5-1 mm above the surface of the ferrofluid. The short and long solenoids were arranged alternatingly due to geometrical considerations. The resulting workspace from the top view is a circle with a diameter of up to ~8 mm.

### III. MATERIALS

#### A. Water-based ferrofluid

Water-based ferrofluid ($FeCl_3@H_2O$) containing iron (III) chloride hexahydrate iron (II) chloride tetrahydrate was used in the experimental study. The surface tension, the density, the viscosity, and the evaporation rate of the utilized ferrofluid are summarized in Table I. The properties of the ferrofluid may vary during the manipulation experiments due to evaporation and absorption when exposed to air. Further details on the preparation of the ferrofluid are given in Appendix B and the characterization of the ferrofluid in Appendix C.

When the ferrofluid was poured into the Petri dish, the ferrofluid forms a meniscus arising from contact with the wall of the dish. This wall effect has been minimized by overfilling the dish with ferrofluid, followed by extracting back some of the ferrofluid until the contact angle of the ferrofluid with the wall is ~90°.

#### B. Objects for manipulation

For the manipulation experiments, different types of non-magnetic objects were used. The system was characterized using a 550 µm diameter polyethylene (PE) spherical particle (PE, Cospheric LLC, USA). Other objects used for the experiments include a 1 mm diameter polystyrene (PS) spherical particle (PSS-1.05, Cospheric LLC, USA) and a 980 µm × 980 µm × 540 µm sized silicon chip. Additionally, poppy and sesame seeds were used for demonstrating the manipulation of biologically relevant matter.

### IV. OPEN-LOOP PERFORMANCE

To examine the trajectory of particle motion under manipulation, a PE particle was placed about 3 mm in front of a short solenoid and then the solenoid was actuated with a constant current of 1.43 A. The resulted trajectories for two repetitions of the open-loop tests are shown in Fig. 2(a), and see the Supplementary multimedia file for video data. The actuation of the solenoid pushes the particle to the center of the workspace and beyond in a slightly curved line attributed to the topology of the ferrofluid surface during manipulation. The motion of the particle is largely repeatable as shown in the figure. During the tests, the particle passed the center and much beyond, showing that the manipulator is capable of manipulating particles within the workspace enclosed by the tip of the solenoids.

To characterize the velocity of particle motion under manipulation, we calculate the mean velocity during the first second after actuation of an initially resting PE particle. The mean velocity during the first second, or actuation velocity hereafter, as a function of the particle-solenoid initial distance was experimentally studied. Fig. 2(b) depicts the mean and standard deviation of the actuation velocity of the actuated particles at five different distances (2.7 mm, 3.8 mm, 4.9 mm, 5.9 mm, 7.0 mm) at constant excitation current of 1.43 A, where the experiment was repeated five times for each distance. The actuation velocity was $1.18 \pm 0.05$ mm/s at 2.7 mm initial distance and monotonically decreasing to $0.49 \pm 0.05$ mm/s at 7.0 mm initial distance. This dependency can be best described as an inverse function of distance $v_a = 3.17\rho^{-1} + 0.03$, where $v_a$ is the actuation velocity of the particle, in mm/s, and $\rho(x,y) = \sqrt{x^2 + y^2}$ is the projected distance on the ferrofluidic surface from the center of the solenoid tip to the particle, in mm. Additionally, negative proportional relation $v_a = -0.16\rho + 1.54$ is given, as well as inverse squared relation $v_a = 6.02\rho^{-2} + 0.40$. The standard deviation of the actuation velocity of the particle is greater when the initial position is closer to the solenoid, attributed to the dynamics of the interface deformation. Additionally, an increase of the standard deviation occurs when the initial position is further away from the solenoid due to the drift.

The influence from the actuation current to the actuation velocity of the particle is studied by repeating the pushing experiments at the same initial position of 3.8 mm at current values of 0.24 A, 0.48 A, 0.72 A, 0.95 A, 1.19 A, 1.43 A, and 1.66 A. The experiment was repeated five times for each current value. Fig. 2(c) shows that the actuation velocity of the particles is linear with respect to the current and can be described as $v_a =$ $0.66I - 0.05$, where $I$ is the actuation current. The standard deviation of the actuation velocity increases with the actuation current, attributed to the greater oscillation of the ferrofluidic interface deformation caused by the greater step actuation current of the solenoid. The drift of the PE particles on the air-ferrofluid interface become increasingly significant compared to the pushing motion of the ferrofluid at farther side of the workspace when the actuation current of the solenoid is below 0.95 A.

TABLE I.    PROPERTIES OF THE UTILIZED FERROLFUID

| Parameter | Value |
|---|---|
| Viscosity (mP·s) | 0.9 |
| Density (kg/m³) | 1071 |
| Surface Tension (mN/m) | 74.75 (±0.15) |
| Evaporation (ml/h) | 0.6 |

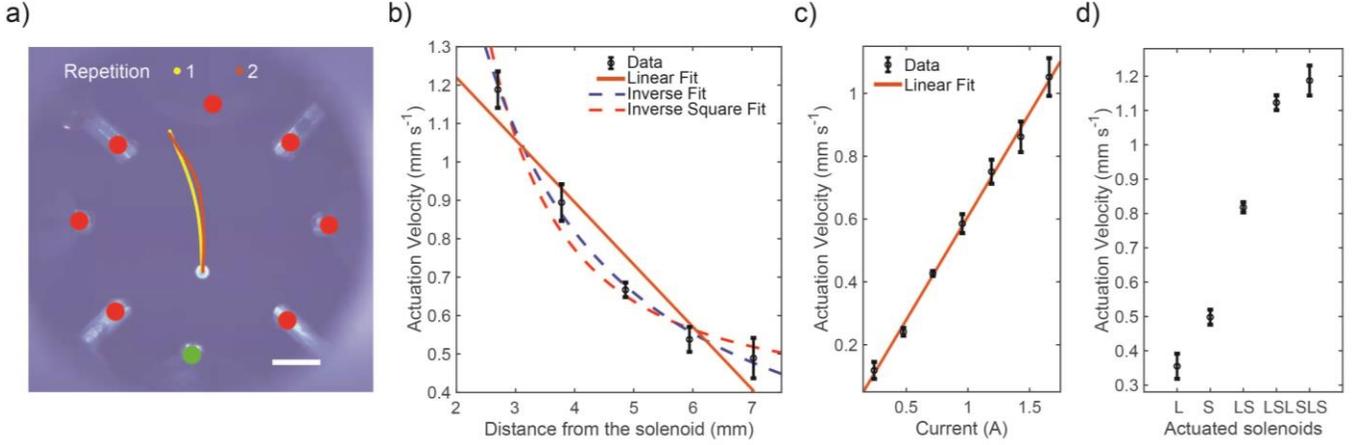

Fig 2. Open-loop performance the ferrofluid manipulator **a)** A polyethylene (PE) particle motion induced by a single solenoid. The scale bar is 2 mm. **b)** Actuation velocity of the PE particle placed at different initial distances from the actuated solenoid; **c)** Actuation velocity of the PE particle when a solenoid is actuated with different currents. **d)** Actuation velocity of the PE particle when a different number of solenoids are actuated simultaneously, where L represents a long solenoid, and S represents a short solenoid.

To study the actuation ability of multiple solenoids, we measured the particle actuation velocity when actuated by one, or simultaneously two, or three solenoids with the initial particle position at the center of the workspace, i.e., 5.1 mm from the solenoid. The results (Fig. 2(d)) show that the actuation velocity of a particle for one actuated long solenoid was $0.36\pm0.04$ mm/s, for one actuated short solenoid was $0.50\pm0.02$ mm/s, for two solenoids where one short and one long was $0.82\pm0.01$ mm/s, for two long solenoid and one short solenoid was $1.12\pm0.02$ mm/s, and for two short solenoids and one long solenoid was $1.19\pm0.04$ mm/s, respectively. The results are only for reference since there are slight variations of the magnetic fields among the solenoids.

## V. AUTOMATIC PATH-FOLLOWING CONTROL

To achieve automatic path-following of the particles, we use a linear programming-based visual servoing control algorithm. The inputs to the algorithm are the current position $P\ (x_p, y_p)$ of the particle and the target location $T\ (x_t, y_t)$ as illustrated in Fig. 3. Both, $P$ and $T$, are defined in respect to the origin $O$ (0, 0), which is in the upper left corner of the frame, not shown in the figure. The current position of the particle is identified by performing grey-level thresholding [35] and blob detection on the images captured by the camera. The output of the algorithm $S$, is a $1 \times 8$ array of zeros and ones, where a zero denotes the OFF state and a one denotes the ON state of a solenoid. The actuation velocity of the particle with respect to the distance from the solenoid has been experimental derived as indicated in Fig. 2(b). Therefore, in real-time trajectory and positioning control, we can estimate the expected motion of the particle using the understanding of the actuation velocity obtained in the previous section. Specifically:

$$\bm{V_{m_{i=1..8}}} = \left(a\|\overrightarrow{MP}\|_i^{-1} + b\right) \cdot \frac{\overrightarrow{MP_i}}{\|\overrightarrow{MP}\|_i} \quad (2)$$

where $\bm{V_{m_{i=1..8}}}$ are expected velocity projections on the particle, $a$ and $b$ are model coefficients and $\overrightarrow{MP}_{i=1..8}$ are the displacement vectors from each solenoid to the micro-object, $\|\cdot\|$ is the vector norm operator.

The scalar projections of $V_m$ parallel and orthogonal to $\overrightarrow{PT}$ is denoted by $V_{m\|\ i=1..8}$ and $V_{m\perp\ i=1..8}$ respectively. This scalar projection can be expressed as the following dot products:

$$V_{m\|\ i=1..8} = \bm{V_{m_{i=1..8}}} \cdot \frac{\overrightarrow{PT}}{\|\overrightarrow{PT}\|}, V_{m\perp\ i=1..8} = \bm{V_{m_{i=1..8}}} \cdot \frac{\overrightarrow{PT_\perp}}{\|\overrightarrow{PT_\perp}\|} \quad (3)$$

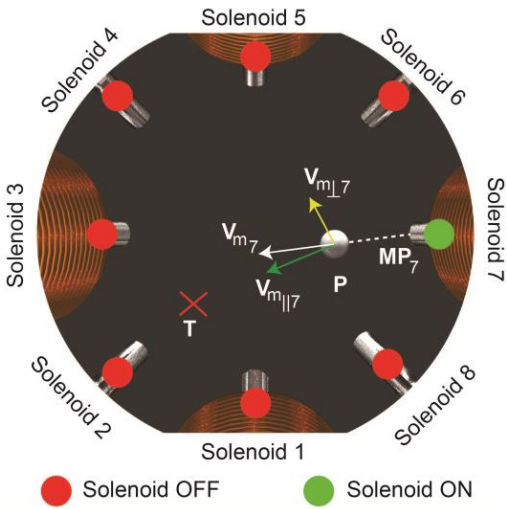

Fig. 3. Illustration of the path-following control algorithm. A particle $P$ in the workspace is defined with its position coordinates $(x_p, y_p)$ and a target $T\ (x_t, y_t)$ in respect to the origin $O$ (upper-left corner of the frame). The particle moves with $\bm{V_{m7}}$ since solenoid 7 is ON. $V_{m\|7}$ and $V_{m\perp 7}$ are parallel and orthogonal projections of $\bm{V_{m7}}$ respectively. The particle is at a distance $MP_7$ from the solenoid 7.

where $\overrightarrow{PT_\perp}$ is the orthogonal vector of $\overrightarrow{PT}$.

The main objective of the control algorithm is to reach the particle to the target by maximizing the motion along $\overrightarrow{PT}$ and minimizing the motion along $\overrightarrow{PT_\perp}$ without any limitation of simultaneous actuation of multiple solenoids. This objective is a multi-objective optimization problem whose cost function $J$ can be represented as:

$$J = \sum_{i=1}^{8} \alpha(V_{m\|i}S_i) - \beta(|V_{m\perp i}|S_i) - \gamma\left(\frac{1}{\|\overrightarrow{PT}\|^2}S_i\right) \quad (4)$$

where $|\cdot|$ is the absolute-value norm. The last term in the cost function is distance-dependent between the particle and the target $\|\overrightarrow{PT}\|$, so that number of actuated solenoids is reduced when the particle is closer to the target and vice versa.

The parameters $\alpha, \beta, \gamma$ were experimentally tuned such that $\alpha = 0.4145$, $\beta = 0.2685$ and $\gamma = 0.0001$. The cost function was solved using the mixed-integer linear programming method (MILP) [36] and a weighted sum approach implemented using CPLEX 12.10 library.

The performance of the control method was evaluated with line, square and circular path-following repeatability tests using a PE particle as shown in Fig. 4, and see the Supplementary multimedia file for videos. All solenoids were excited with a constant voltage of 6 V and a corresponding current of about 1.4 A. The trajectories were repeated ten times where only three repetitions are shown in Fig. 4 for clarity. The particle was moved from left to right following a 4 mm line where the origin of the plot is at the center of the workspace. The relatively greater deviation from the reference trajectory at the starting position can be attributed to the step actuation of the ferrofluid. At the end of the trajectory, the deviation is also greater due to the greater distance between the particle to the solenoids in actuation (see Fig. 2(b)).

Table II summarizes the path following errors for all three paths. Similar to the line path, greater deviation of the square path also occurs at the starting position due to step actuation, and the increased distance from the actuation solenoids to the particle. However, the square path also involves the particle motion in the off-center path, where the distance to actuation solenoids is asymmetric, resulting in slightly greater standard deviation and maximum error.

Table III summarizes the resultant velocities when manipulating the particle in the reference trajectories. For the line following, the average velocity is the lowest. The reason is that the motion is induced only by the solenoids on the left side (solenoid 2, 3 and 4), which was initially effective but weakens with the increasing distance between the actuated solenoids and the particle, especially at the end of the trajectory. On the other hand, the average velocity for circular path following is the highest, since the average distances between the actuation solenoid and the particle are relatively closer along the path. A greater deviation from the reference paths in the case of square and circular trajectories were also observed at the top side near solenoid 4, 5 and 6. These deviations can be attributed to the differences among the solenoids.

Additionally, a PE particle has been brought to the center of the workspace and its position has been maintained by the control algorithm for a duration of 60 seconds. The experiment has been performed at 3 different actuation currents i.e. 0.95, 1.19 and 1.43 A. The particle remained in the targeted position with a positioning error of $34.3 \pm 18.1$ µm, $60.71 \pm 30.7$ µm and $76.5 \pm 35.7$ µm, respectively for each current. The errors translate to ~6.2 %, 11.0 % and 13.9 % of the particle body length.

TABLE II. PATH FOLLOWING ERRORS

| Path | Mean Error (µm) | Standard Deviation (µm) | Max Error (µm) |
|---|---|---|---|
| Line | 25.1 | 19.5 | 104.3 |
| Square | 34.4 | 28.4 | 214.8 |
| Circle | 33.4 | 26.6 | 192.9 |

TABLE III. VELOCITY DURING PATH FOLLOWING

| Path | Mean Velocity (µm/s) | Standard Deviation (µm/s) | Max Velocity (µm/s) |
|---|---|---|---|
| Line | 270.0 | 52.7 | 341.9 |
| Square | 440.7 | 33.4 | 467.8 |
| Circle | 509.5 | 43.7 | 592.8 |

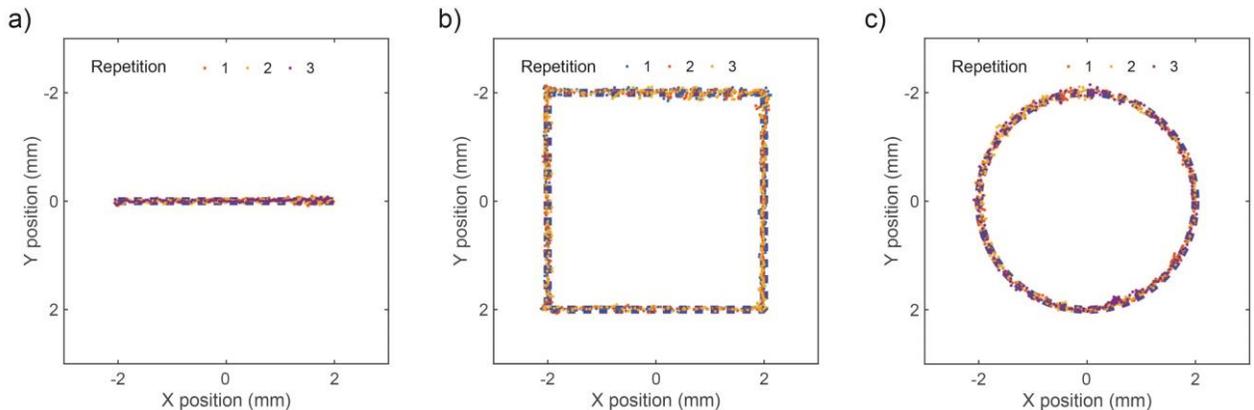

Fig 4. Automatic path-following control of a PE particle for reference paths of **a)** line **b)** square, and **c)** circle. The blue dashed line is the reference trajectory.

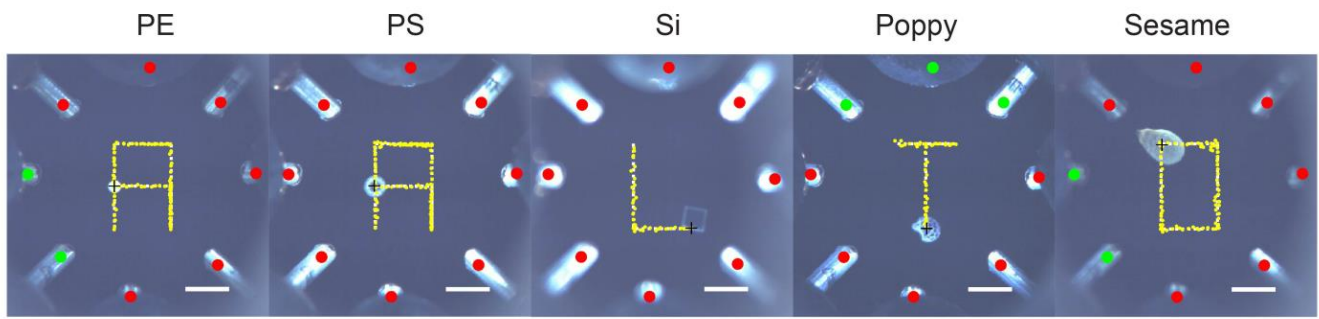

Fig 5. Automatic path-following manipulation of a polyethylene (PE) and a polystyrene (PS) particle, a silicon (Si) chip; a poppy seed and sesame in writing the letters "AALTO". The trajectory (in yellow) is down sampled three times. The scale bar is 2 mm.

Fig. 5 shows the manipulation of different types of objects in writing the letters "AALTO". A PE particle and a PS particle were manipulated in writing the letter "A". A silicon chip was manipulated in writing the letter "L". Poppy and sesame seeds were manipulated in writing the letters "T" and "O". The proposed algorithm is adaptable in manipulating objects of different materials, shapes, and sizes merely by altering the supply voltage/current. The polyethylene (PE) spherical particle was manipulated with 6 V supply voltage, translating to ~1.4 A in current. The polystyrene (PS) spherical particle was manipulated with 5 V, or ~1.2 A. The silicon chip is manipulated with 2 V, or ~0.47 A. The poppy and sesame seeds are manipulated with 4 V and 3 V, or 0.95 and 0.7 A, respectively. The actuation current (or voltage) was adjusted based on the type of the particle to achieve a comparable velocity. See the Supplementary multimedia file for videos.

## VI. SUMMARY AND DISCUSSION

This paper reported a ferrofluid manipulator that can automatically manipulate floating non-magnetic micro-objects by deforming the interface of a ferrofluid using eight centimeter-scale solenoids. Employing a linear programming-based control algorithm, the manipulator can control the motion of non-magnetic particles following different paths, from a line to different letter-like trajectories. The achieved path following precision is as good as 25 µm for a 0.55 mm PE particle. For achieving more precise path-following control, a pulse-width modulation (PWM) or current control circuits for each solenoid could be constructed. Besides the simplicity of the path following control algorithm, the manipulator can manipulate non-magnetic particles of different materials and of different shapes, including spherical polyethylene (PE) and polystyrene (PS) particles, a square silicon (Si) chip, and natural shaped poppy and sesame seeds.

The utilization of eight solenoids renders this ferrofluidic manipulator an over-actuated system. In principle, three solenoids are enough to induce controlled planar motion, but in such a case, the workspace will be a triangle. From a general perspective, the effective workspace is defined by the polygon with a solenoid as a vertex. More solenoids correspond to a wider workspace and the ability to simultaneously actuate multiple solenoids expand the size of the workspace. The ferrofluidic manipulator was designed to be over-actuated so the automated motion of the particles is smoother, and the size of the workspace is sufficient.

This micromanipulation approach can be found useful in combination with biocompatible ferrofluids for the manipulation of cell cultures and biological tissues that must be grown on the air-liquid interface. Such interfacial bio-systems can be used to mimic respiratory tract in *in-vitro* conditions [37].

In future work, we aim to provide a complete mathematical description of particle motion in the ferrofluidic manipulator. Further, the proposed technique could be extended to using biocompatible ferrofluids [38] for the manipulation of cellular cultures on the air-ferrofluid interface. Some potential modifications employing electromagnetic needles [12] [13] could also be considered to concentrate the magnetic field for manipulating smaller objects as well as to decrease the impact of possible geometrical constraints. Another direction that could be explored is on the manipulation of multiple objects on the ferrofluidic interface by employing machine learning-based approaches.

## VII. APPENDIX A: DESCRIPTION OF THE MECHATRONIC SYSTEM

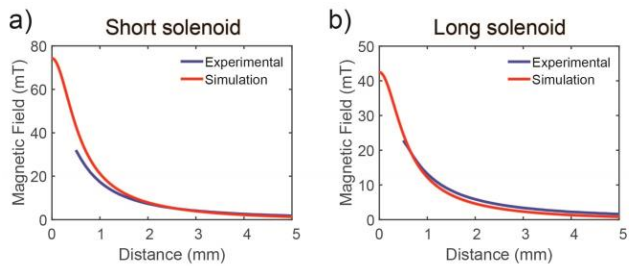

Fig. A1. Magnetic field intensity over distance of **a)** a short and b) a long solenoid excited with 1,15 A current (or voltage of 5V).

### A. Magnetic characteristics of the solenoids

The magnetic field intensity of a short (Fig. A1(a)) and a long solenoid (Fig. A1(b)) at 1.15 A excitation current was experimentally measured using a hall sensor (SS495A1, Honeywell Inc., USA). The numerical simulation of the magnetic field conducted in Comsol Multiphysics 5.2a (Comsol Group, Sweden), where the two are in good agreement.

### B. Electronic circuitry

The solenoids were actuated by a custom-made electronic circuitry (Fig. A2). Each coil is serially connected with a power resistor $R_{PR} = 3.3\ \Omega$ (Vishay, USA), and in parallel connection with fly-back diode 1N4003 (Multicomp, Farnell, UK), and a 3 mm LED diode (Multicomp, Farnell, UK) and 330 Ω resistor (TE Connectivity, Switzerland) for debugging. The coil is switched by a solid-state relay (Finder, Italy). This sub-circuitry has been used for each coil, powered by a laboratory power supply (DC E3634A, Keysight, USA). A 0.1 Ω shunt resistor with ≥ 3 W power rating (TT Electronics, UK) was installed to monitor the current consumption. The relays were controlled by a digital input/output board (NI DIO 6501, National Instruments, USA), where each output channel was connected in parallel to a 4.7 KΩ pull-up resistor (TT Electronics, UK). The overall resistance in one line in which a solenoid is actuated is ~4.2 Ω (3.3 Ω from a solenoid, 0.5 Ω from the power resistor, and 0.4 Ω from resistance in the cables). The current-voltage dependency in each solenoid was heuristically identified as linear for the voltages range in the experiments, where 2–6 V corresponds to 0.47 A – 1.4 A. Due to the minor variation of the resistances among different solenoids, the constant voltage from the power supply may lead to slightly different currents in each solenoid.

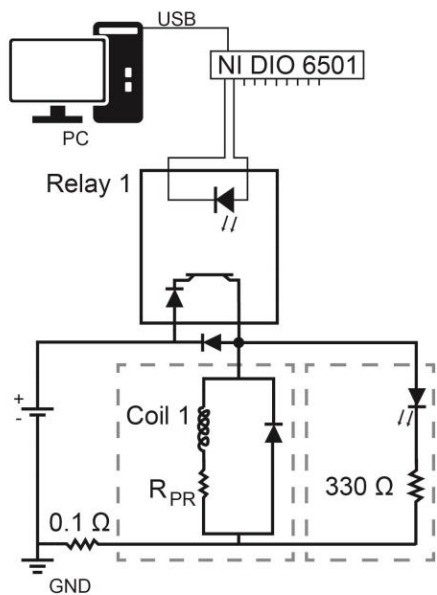

Fig. A2. Simplified schematic of the custom-made electronic circuitry utilized in the ferrofluidic manipulator. Only one out of eight coils with a relays and accompanying resistors and diodes is illustrated. Other missing components in the illustration are the pull-up resistors within the NI DIO 6501, the LEDs, and resistors for the LEDs on the Relay's input.

## VIII. APPENDIX B. PREPARATION OF WATER-BASED FERROFLUID

Water-based ferrofluid (FeCl$_3$@H$_2$O) was prepared through the coprecipitation method. 8.65 g of iron (III) chloride hexahydrate and 3.18 g of iron (II) chloride tetrahydrate were dissolved in 288 g of MQ water followed by coprecipitation through the addition of 32 ml of ammonium hydroxide (28 % – 30 %). The resulting black dispersion was stirred for 5 min after which 10.1 g of citric acid monohydrate was added to stabilize the particles followed by another 5 min of stirring. The particles were sedimented with a strong magnet, aqueous supernatant discarded by decantation, and then re-dispersed with an additional 1.86 g of citric acid monohydrate in 80 ml of MQ H$_2$O by stirring for 5 min. 160 ml of acetone was added, the dispersion was then magnetically decanted, and the particles were redispersed in 80 ml of MQ H$_2$O. This magnetic decantation washing cycle was repeated twice. After one more cycle of magnetic decantation, the particles were re-dispersed in MQ H$_2$O to obtain a water-based ferrofluid.

## IX. APPENDIX C: CHARACTERIZATION OF UTILIZED MATERIALS

The magnetic moment of the ferrofluids was measured in a multipurpose measurement system (PPMS Dynacool, Quantum

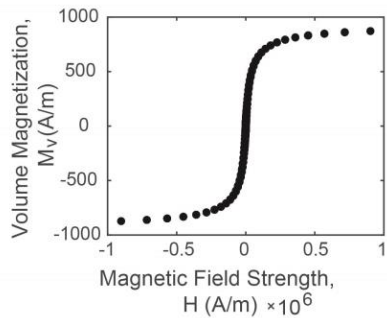

Fig. A3. Magnetic characteristics of a ferrofluid with a concentration of 12.4 mg/ml of iron (III) chloride in water as a carrier medium.

Design, USA) using the Vibrating Sample Magnetometer (VSM) mode with a range from -1 to 1 T. Fig. A3 shows the magnetic characteristics of the ferrofluids used in our experimental work. One should note that the nanoparticle concentration of the water-based ferrofluid increased slowly with time due to evaporation, since the ferrofluid was reused in several experiments.

The density of the ferrofluid was calculated as ratio between the mass of a ferrofluid and the volume (1 mL) of ferrofluid filled in a syringe with 1 mL capacity (Hamilton Company, Switzerland). The surface tension of a ferrofluid sample was measured by contact angle goniometer Attension® Theta (Biolin Scientific, Sweden). The viscosity of the ferrofluid was measured with a rheometer Physica MCR 300 (Anton Paar GmbH, Austria). The evaporation rate was measured by observing the Petri dish (diameter: 55 mm) filled with ferrofluid and measuring its mass with a precision scale (EK-400H, A&D Company LTD, Japan) every 10 minutes for overall duration of 90 to 120 minutes per sample.

## ACKNOWLEDGMENT

The authors would also like to express their gratitude to Heikki Hyyti, Vesa Korhonen, and Visa Lukkarinen for support in the design and the manufacture of the custom-made electronic board. Z.C. acknowledges the incentive stipend provided by Neles Foundation.

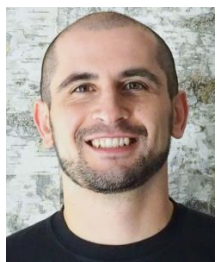

**Zoran Cenev** received his B.Sc. degree in mechatronics from Saints Cyril and Methodius University, Skopje, Macedonia, in 2011 and the M.Sc. degree in mechatronics and micromachines from the Tampere University of Technology (current name: Tampere University), Tampere, Finland, in 2014. He completed his Ph.D. degree in micro- and nanorobotics under supervision of Prof. Zhou at the Aalto University, Espoo, Finland, in 2019. He is currently a postdoctoral researcher in active matter under supervision of Prof. Timonen at the same university and his research interests include magnetic-based micromanipulation, phenomena at fluid–fluid interfaces, levitodynamics, active matter, and wearable electronics.

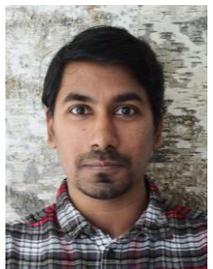

**P. A. Diluka Harischandra** received his B.Sc. degree in Electronics Engineering and M.Eng. degree in Mechatronics Engineering from Asian Institute of Technology, Thailand in 2015 and 2017. He is currently pursuing the Ph.D. degree with the Robotic Instruments Group, Aalto University, Finland, where he has been developing automatic control methods for magnetic manipulation of matter at air-liquid interfaces. His research interests include micromanipulation, automation, visual servo control, haptics, and intelligent robotics.

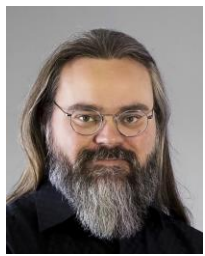

**Mika Latikka** received the M.Sc. degree in physics from University of Helsinki, Finland, in 2014, and the Ph.D. degree in engineering physics from Aalto University, Finland, in 2020. He is currently a postdoctoral researcher in Soft Matter and Wetting research group lead by Professor Robin Ras in Aalto University, developing advanced wetting characterization instruments based on magnetic droplets. His research interest include wetting, superhydrophobicity, magnetic nanoparticles and ferrofluids.

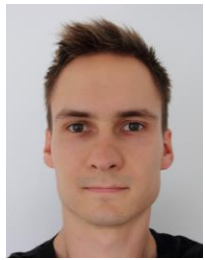

**Ville Hynninen** received his B.Sc. (2012) in biochemistry, and M.Sc. (2014) in molecular biology from Tampere University (Finland). He graduated as a D.Sc. (Tech.) in the field of engineering physics from Aalto University (Finland) under the supervision of Professor Olli Ikkala in 2021. Currently, he works as a joint postdoctoral researcher between Aalto University and Tampere University focusing on biopolymer-based nanocomposite fibers and optical materials.

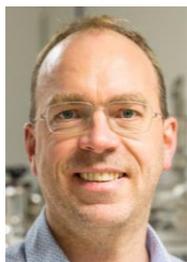

**Robin H. A. Ras** is associate professor at Aalto University School of Science, and Head of Department of Applied Physics. He is the recipient of prestigious European and Finnish grants, including ERC Consolidator Grant, ERC Proof-of-Concept, Academy of Finland, Business Finland (TUTL and R2B), Future Makers. His work has been published in high-visibility journals, including Science, Nature, Advanced Materials, Advanced Science, PNAS, Science Advances. Prof. Ras won the Anton Paar Research Award for Instrumental Analytics and Characterization in 2018. His main research interest is in wetting/non-wetting surfaces, wetting metrology, fluorescent metal nanoclusters, responsive polymers, nanocomposites, aerogels. His research is partly inspired by materials from the biological world (biomimetics).

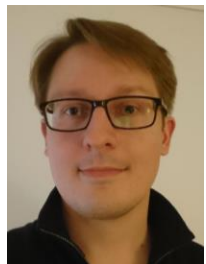

**Jaakko Timonen** is an assistant professor at Aalto University School of Science, Department of Applied Physics. He is also a Research Fellow of the Academy of Finland (2018-2023), a recipient of the ERC Starting Grant (2019-2023), and a member of the Young Academy Finland (YAF). His main research interest is in experimental soft condensed matter physics, with emphasis on various capillary phenomena, colloidal systems, active matter, and emergence of complex behavior in apparently simple systems when driven out of the thermodynamic equilibrium.

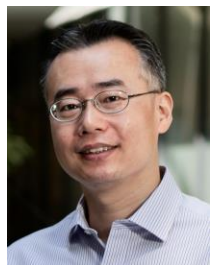

**Quan Zhou** received the M.Sc. and Dr.Tech. degrees from the Tampere University of Technology, Finland, in 1996 and 2004, respectively. He is currently an associate professor of automation leading the Robotic Instruments Group, School of Electrical Engineering, Aalto University, Finland. He was a professor with the Northwest Polytechnical University, Xi'An, China. His main research interests include miniaturised robotics, micro and nanomanipulation, and automation methods. He is currently the Coordinator of Topic Group Miniaturised Robotics of European Robotics Association (euRobotics). He was also the Coordinator of the EU FP7 Project FAB2ASM. He was the General Chair of the International Conference on Manipulation, Automation, and Robotics at Small Scales (MARSS) in 2019, and the Chair of the IEEE Finland Joint Chapter of Control System Society, Robotics, and Automation Society, and System Man and Cybernetics Society, from 2015 to 2016. He has also won the 2018 Anton Paar Research Award for Instrumental Analytics and Characterization.